\documentclass[sn-mathphys-num]{sn-jnl}% Math and Physical Sciences Numbered Reference Style 
\usepackage{graphicx}%
\usepackage{multirow}%
\usepackage{amsmath,amssymb,amsfonts}%
\usepackage{amsthm}%
\usepackage{mathrsfs}%
\usepackage[title]{appendix}%
\usepackage{xcolor}%
\usepackage{textcomp}%
\usepackage{manyfoot}%
\usepackage{booktabs}%
\usepackage{algorithm}%
\usepackage{algorithmicx}%
\usepackage{algpseudocode}%
\usepackage{listings}%
\usepackage{soul}
\usepackage[most]{tcolorbox}
\usepackage{array}
\newcolumntype{L}[1]{>{\raggedright\let\newline\\\arraybackslash\hspace{0pt}}m{#1}}
\newcolumntype{C}[1]{>{\centering\let\newline\\\arraybackslash\hspace{0pt}}m{#1}}
\newcolumntype{R}[1]{>{\raggedleft\let\newline\\\arraybackslash\hspace{0pt}}m{#1}} 

\usepackage{makecell}

\theoremstyle{thmstyleone}%
\theoremstyle{thmstyletwo}%

\theoremstyle{thmstylethree}%

\usepackage[numbers]{natbib}
\usepackage{xcolor}

\newcommand\highlightReference[1]{%
  \expandafter\newcommand\csname highlightReference-#1\endcsname{}%
}

\usepackage{xpatch}
\makeatletter
\xapptocmd\@lbibitem{\hlbibitem{#2}}{}{}
\makeatother
\def\hlbibitem#1 #2\par{%
  \expandafter\ifx\csname highlightReference-#1\endcsname\relax
    #2\par
  \else
    \textcolor{red}{#2}\par
  \fi
}

\highlightReference{dohmatob2024tale}
\highlightReference{ferbach2024self}
\highlightReference{feng2024beyond}
\highlightReference{gerstgrasser2024model}
\highlightReference{fu2024towards}
\highlightReference{gillman2024self}
\highlightReference{guo2023curious}
\highlightReference{dohmatob2024model}
\highlightReference{bertrand2023stability}
\highlightReference{RN5}
\highlightReference{RN11}
\highlightReference{RN12}
\highlightReference{bohacek2023nepotistically}
\highlightReference{RN4}
\highlightReference{briesch2023large}
\highlightReference{alemohammad2024self}

\begin{document}

\title[AI Autophagy]{On the Caveats of AI Autophagy}

%%=============================================================%%
%% GivenName	-> \fnm{Joergen W.}
%% Particle	-> \spfx{van der} -> surname prefix
%% FamilyName	-> \sur{Ploeg}
%% Suffix	-> \sfx{IV}
%% \author*[1,2]{\fnm{Joergen W.} \spfx{van der} \sur{Ploeg} 
%%  \sfx{IV}}\email{iauthor@gmail.com}
%%=============================================================%%

%\author*[1,2]{\fnm{First} \sur{Author}}\email{iauthor@gmail.com}

%\author[2,3]{\fnm{Second} \sur{Author}}\email{iiauthor@gmail.com}
%\equalcont{These authors contributed equally to this work.}

%\author[1,2]{\fnm{Third} \sur{Author}}\email{iiiauthor@gmail.com}
%\equalcont{These authors contributed equally to this work.}

%\affil*[1]{\orgdiv{Department}, \orgname{Organization}, \orgaddress{\street{Street}, \city{City}, \postcode{100190}, \state{State}, \country{Country}}}

%\affil[2]{\orgdiv{Department}, \orgname{Organization}, \orgaddress{\street{Street}, \city{City}, \postcode{10587}, \state{State}, \country{Country}}}

%\affil[3]{\orgdiv{Department}, \orgname{Organization}, \orgaddress{\street{Street}, \city{City}, \postcode{610101}, \state{State}, \country{Country}}}

\author[1,2]{\fnm{Xiaodan} \sur{Xing}}

\author[1]{\fnm{Fadong} \sur{Shi}}
\author[1,2]{\fnm{Jiahao} \sur{Huang}}
\author[1,2]{\fnm{Yinzhe} \sur{Wu}}
\author[1,2]{\fnm{Yang} \sur{Nan}}
\author[1,2]{\fnm{Sheng} \sur{Zhang}}

\author[1,2]{\fnm{Yingying} \sur{Fang}}
\author[5,6]{\fnm{Michael} \sur{Roberts}}
% \author[1,2,3]{\fnm{Guang} \sur{Yang}}

\author[5]{\fnm{Carola-Bibiane} \sur{Schönlieb}}

\author[7,8]{\fnm{Javier} \sur{Del Ser}}
\author[1,2,3,4]{\fnm{Guang} \sur{Yang}}

\affil*[1]{\orgdiv{Bioengineering Department and Imperial-X}, \orgname{Imperial College London}, \orgaddress{
%\street{Exhibition Road}, \city{London}, \postcode{SW7 2AZ}, 
\state{London}, \country{UK}}}

\affil[2]{\orgdiv{National Heart and Lung Institute}, \orgname{Imperial College London}, \orgaddress{
%\street{Dovehouse Street}, \city{London}, \postcode{SW3 6LY}, 
\state{London}, \country{UK}}}

\affil[3]{\orgdiv{Cardiovascular Research Centre}, \orgname{Royal Brompton Hospital}, \orgaddress{
%\street{Sydney Street}, \city{London}, \postcode{SW3 6NP}, 
\state{London}, \country{UK}}}

\affil[4]{\orgdiv{School of Biomedical Engineering \& Imaging Sciences, \orgname{King's College London}, \orgaddress{
%\street{Strand}, \city{London}, \postcode{WC2R 2LS}, 
\state{London}, \country{UK}}}}

\affil[5]{\orgdiv{Department of Applied Mathematics and Theoretical Physics}, \orgname{University of Cambridge}, \orgaddress{
%\street{Wilberforce Road}, \city{Cambridge}, \postcode{CB3 0WA}, 
\state{Cambridge}, \country{UK}}}

\affil[6]{\orgdiv{Department of Medicine}, \orgname{University of Cambridge}, \orgaddress{
%\street{Wilberforce Road}, \city{Cambridge}, \postcode{CB3 0WA}, 
\state{Cambridge}, \country{UK}}}

\affil[7]{\orgdiv{Department of Communications Engineering}, \orgname{University of the Basque Country UPV/EHU}, \orgaddress{
%\street{Alda. Urquijo s/n}, \city{Bilbao}, \postcode{48013}, 
\state{Bilbao}, \country{Spain}}}

\affil[8]{\orgname{TECNALIA, Basque Research and Technology Alliance (BRTA)}, \orgaddress{
%\street{Moneybaixo 700}, \city{Derio}, \postcode{48160}, 
\state{Bilbao}, \country{Spain}}}

%%==================================%%
%% Sample for unstructured abstract %%
%%==================================%%

\abstract{Generative Artificial Intelligence (AI) technologies and large models are producing realistic outputs across various domains, such as images, text, speech, and music. Creating these advanced generative models requires significant resources, particularly large and high-quality datasets. To minimise training expenses, many algorithm developers use data created by the models themselves as a cost-effective training solution. However, not all synthetic data effectively improve model performance, necessitating a strategic balance in the use of real versus synthetic data to optimise outcomes. Currently, the previously well-controlled integration of real and synthetic data is becoming uncontrollable. The widespread and unregulated dissemination of synthetic data online leads to the contamination of datasets traditionally compiled through web scraping, now mixed with unlabeled synthetic data. This trend, known as the AI autophagy phenomenon, suggests a future where generative AI systems may increasingly consume their own outputs without discernment, raising concerns about model performance, reliability, and ethical implications. What will happen if generative AI continuously consumes itself without discernment? What measures can we take to mitigate the potential adverse effects? To address these research questions, this study examines the existing literature, delving into the consequences of AI autophagy, analyzing the associated risks, and exploring strategies to mitigate its impact. Our aim is to provide a comprehensive perspective on this phenomenon advocating for a balanced approach that promotes the sustainable development of generative AI technologies in the era of large models.
}


%%\pacs[JEL Classification]{D8, H51}

%%\pacs[MSC Classification]{35A01, 65L10, 65L12, 65L20, 65L70}

\maketitle

\section{Introduction}\label{sec1}

Deep learning-based generative models have significantly transformed AI and machine learning. These models can generate highly realistic data, with applications in creating images, text, speech, and music. Innovations like DALL-E 2 \cite{dalle2}, Imagen \cite{imagen}, Suno \cite{suno}, and Sora \cite{sora}, alongside advanced language models like ChatGPT \cite{chatgpt}, demonstrate their vast potential in diverse fields such as art, entertainment, medical diagnostics, and automated content generation. These technologies enable seamless integration of visual, textual, and auditory data, showcasing their broad applicability.

Training these deep generative models is not a cheap task. They require extensive, high-quality datasets, leading many organisesations to scrape web data. For instance, CommonCrawl \cite{RN1}, a petabyte-scale open-source web crawling database, is widely used to train generative models such as StableDiffusion and GPT-3. This dataset is regularly expanded with 3–5 billion new pages each month, accounting so far for a total of 250 billion website pages.

As web scraping increases, so does the risk of incorporating synthetic data generated by other models. This means models could be trained on data produced by previous generative models. Currently, most generative models are trained on human-generated content. However, with more synthetic data online, these models may start training on artificial rather than original data. Additionally, the rapid proliferation of data-intensive models is depleting high-quality, human-generated data. Projections suggest that by 2026, the pool of suitable language data for training large language models (LLMs) may be depleted \cite{RN2}. As AI-generated outputs grow, including AI-generated content in training datasets seems inevitable.

Initially, training generative AI on synthetic content might seem beneficial, as integrating the right amount can yield impressive outcomes \cite{RN3}. However, as models evolve over generations, researchers have flagged potential negative impacts \cite{RN4, RN5, RN6}. This phenomenon, where generative models are trained on AI-generated content, is termed "autophagy" \cite{RN5}, similar to the biological process where a cell consumes itself.

% Evidence of these impacts includes the performance degradation of GPT-3.5 and GPT-4 \cite{RN7}, which showed increased formatting errors in code generation from March to June 2023. While the exact causes are unclear, it is speculated that pollution of the training dataset with AI-generated data is a contributing factor \cite{sanusi2023}.

Performance degradation due to synthetic data is a concern beyond generative models. In computer vision, tasks like classification and segmentation can suffer when using synthetic data. Literature shows that fully synthetic data can degrade performance compared to models trained on real data \cite{RN6,hataya2023will}. As models consume synthetic online content, potential consequences in practical settings, especially high-risk scenarios, are significant \cite{RN8}.

\textcolor{black}{Despite the wide recognition of AI autophagy in recent times, a significant gap remains in comprehensive studies  that explore both theoretical and empirical analyses of this phenomenon. Albeit widely discussed in the literature, findings often conflict, leading to divided opinions on the impact of synthetic data on AI models. Some studies suggest that synthetic data pollution might not pose a major concern under controlled conditions: the models' performance will stop deteriorate if carefully managed \cite{bertrand2023stability, gerstgrasser2024model}. In contrast, others present experimental evidence proving the severe and unavoidable consequences of AI autophagy, emphasizing the potential performance deterioration of the model resulting therefrom \cite{RN5, RN11, RN12, briesch2023large}.}

\textcolor{black}{Moreover, there is a clear lack of practical analyses that integrate both technical and regulatory viewpoints. Most existing research tends to focus on isolated technical solutions \cite{feng2024beyond,ferbach2024self}, overlooking the need for a comprehensive approach that examines how these strategies can be effectively implemented in real-world scenarios, particularly within existing worldwide regulatory frameworks.}

\begin{figure}
    \centering
    \includegraphics[width=1\linewidth]{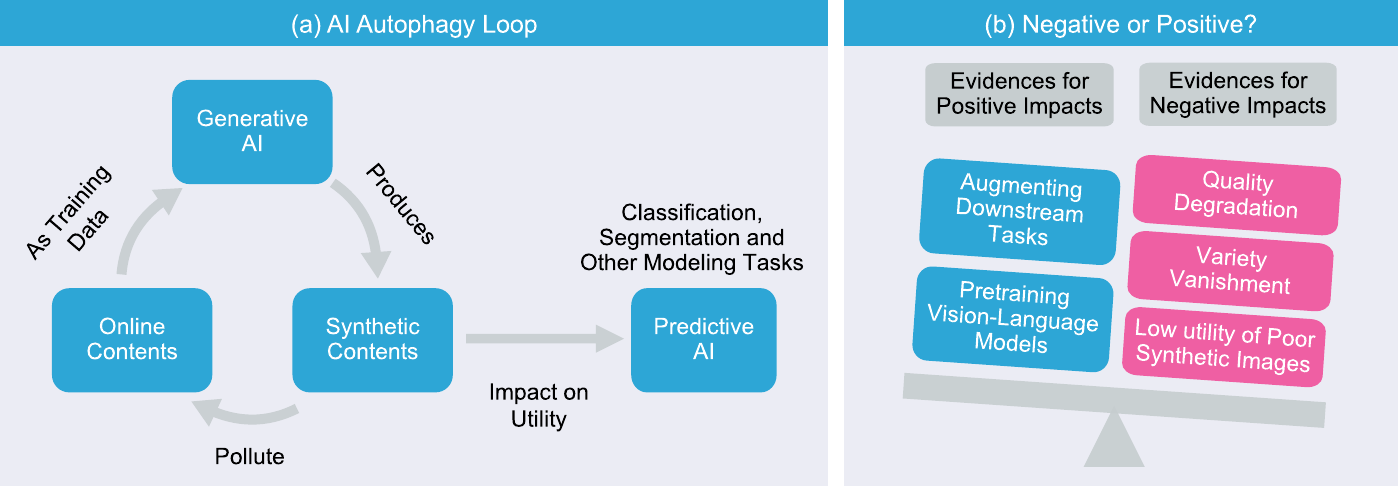}
    \caption{A diagram showing (a) the scope of this work and (b) the major research questions. This study centers on the AI autophagy phenomenon, with a focus on analyzing its effects and discussing solutions to mitigate potential negative impacts. }
    \label{fig:teaser}
\end{figure}

As shown in Figure \ref{fig:teaser}, this work aims to fill the research gap by providing an integrated study that critically examines empirical, theoretical, technical, and regulatory evidence, offering definitive insights and practical recommendations for managing synthetic data in AI development. The objective of this paper is to address three research questions:
\begin{itemize}
    \item \textbf{RQ1:} What are the consequences of out-of-control autophagy with synthetic contents?
    \item \textbf{RQ2:} What technical strategies can be employed to mitigate the negative consequences of AI autophagy?
    \item \textbf{RQ3:} Which regulatory strategies can be employed to address these negative consequences?
\end{itemize}

% The structure of this paper is organized as follows: Section 2 examines the consequences of uncontrolled synthetic content incorporation (RQ1). Section 3 discusses technical strategies to mitigate AI autophagy (RQ2). Finally, Section 4 reviews current regulatory and policy measures to address these negative consequences (RQ3). This approach ensures a comprehensive understanding of the problem and solutions.

\section{RQ1: What Happens When AI Eats Itself?}
The incorporation of synthetic data into training sets is inevitable \cite{RN2} and may reshape our understanding of AI's capabilities and limitations. In this section, we discuss the potential consequences of this trend, specifically examining whether models improve or deteriorate when synthetic data is recursively added to the training dataset. 

\subsection{Fully Synthetic Loop: Worst Case and Theoretical Model}
\textcolor{black}{In the worst-case scenario, the generative model is trained predominantly on its own outputs. This situation is known to distort the data distribution, resulting in a marked decline in model performance \cite{bohacek2023nepotistically,RN4,RN5}. While a fully synthetic loop is not typical in real-world applications, examining this scenario sheds light on the theoretical understanding of \textbf{model collapse}—a phenomenon where AI autophagy leads to severe performance degradation.}

A simple and intuitive way to understand the consequences of a fully synthetic loop is through a Gaussian distribution. The generative process involves sampling data points from a reference Gaussian distribution and using these samples to approximate the original distribution's characteristics. In a fully synthetic loop, each generative model only has access to data generated by previous models. 
\begin{figure}[h]
    \centering
    \includegraphics[width=0.8\linewidth]{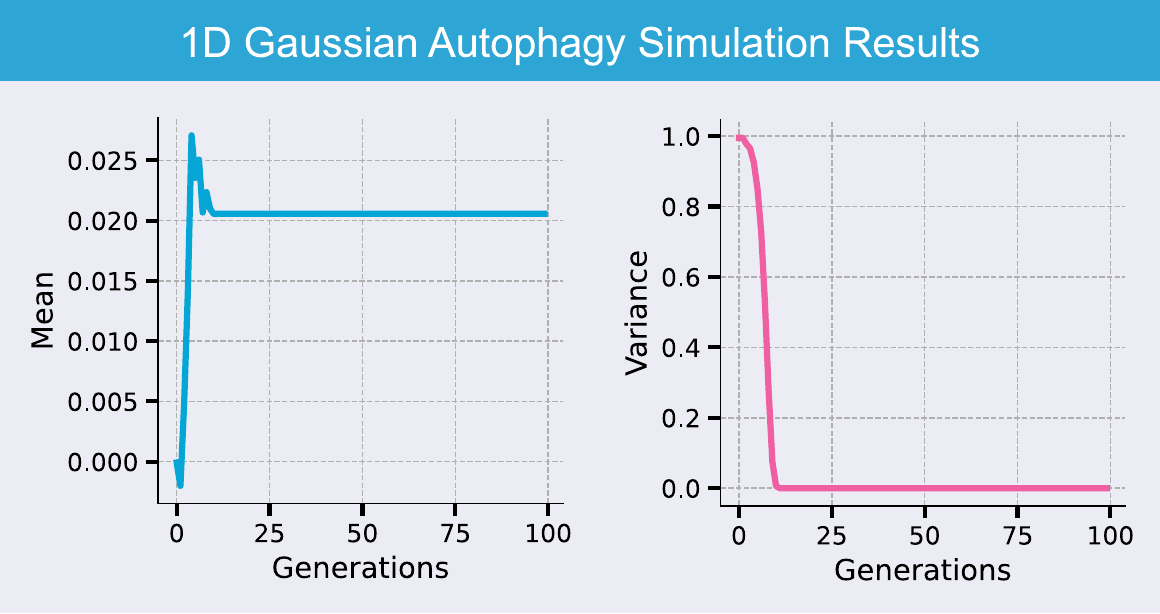}
    \caption{The trend of mean and variance estimations from the autophagy loop. Here we chose the starting value $\mu=0$ and $\sigma=1$. For each round, $10^4$ data points were generated to perform the estimation. }
    \label{fig:1dresults}
\end{figure}

Through this iterative process, as shown theoretically in \cite{RN4, RN5} and in the experimental results in Figure \ref{fig:1dresults}, we observe how the mean and variance evolve across multiple rounds of autophagy: (1) the mean values undergo a random walk, signifying potential degradation in data quality, and (2) the variance decreases over time, leading to a loss of diversity in the generated outputs.

\textcolor{black}{Based on this simple model, Bertrand et al. \cite{bertrand2023stability} extended the analysis to multivariate Gaussian distributions, finding that larger sample sizes in each iteration slow the rate of collapse.}

\textcolor{black}{To investigate how errors propagate through the model's predictions over time, Dohmatob et al. \cite{dohmatob2024model} employed a linear regression model, which offers a clear framework for understanding how errors introduced in each iteration of training can compound over time, leading to significant deviations in the model's predictions. In this setup, they discovered that the test error grows linearly with the number of iterations in fully synthetic loop.}

\begin{figure}
    \centering
    \includegraphics[width=\linewidth]{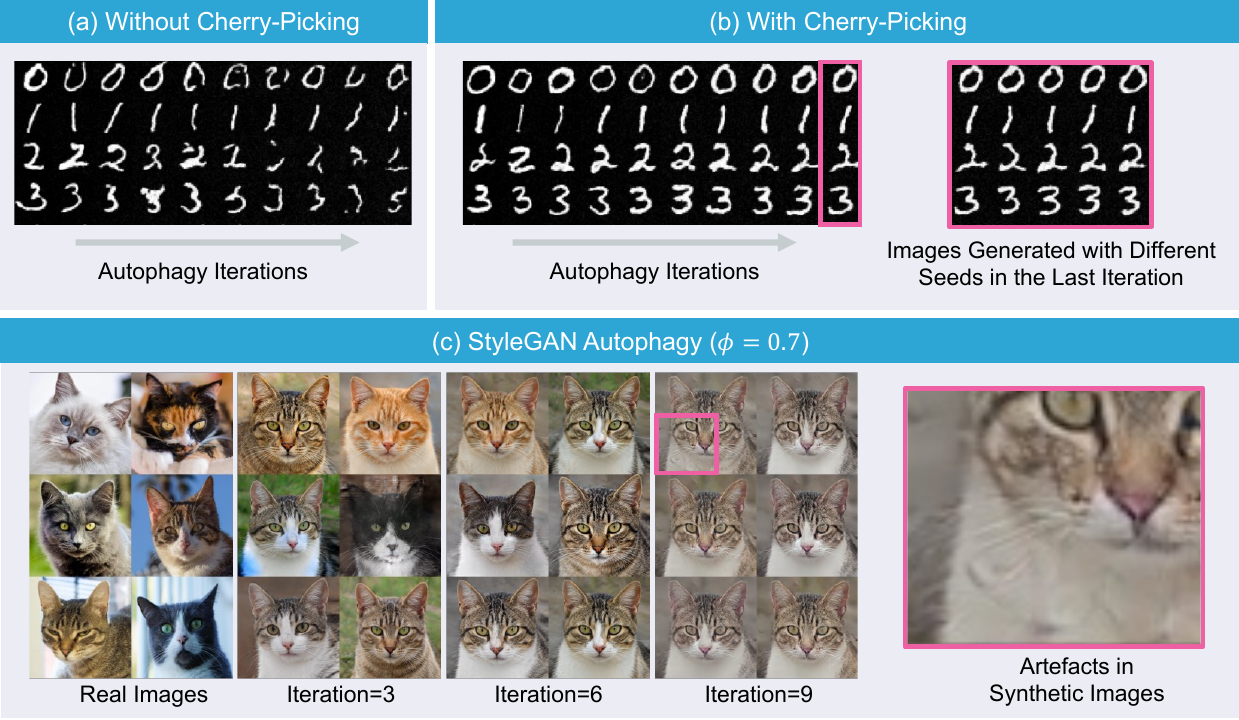}
    \caption{Synthetic images generated from an autophagy loop example using a DDPM model on the MNIST dataset \cite{deng2012mnist} illustrate quality degradation (a). In (b), applying a selective process that retains only high-quality images for subsequent autophagy loops reduces diversity. (c) presents results from a more complex task using the StyleGAN model \cite{karras2020analyzing} on the AFHQ cat dataset  \cite{choi2020stargan}, with a truncation parameter of 0.7 to balance diversity and quality. Despite these adjustments, quality degradation and diversity loss remain evident during autophagy
    }
    \label{fig:vision_quality}
\end{figure}

\begin{figure}
    \centering
    \includegraphics[width=1\linewidth]{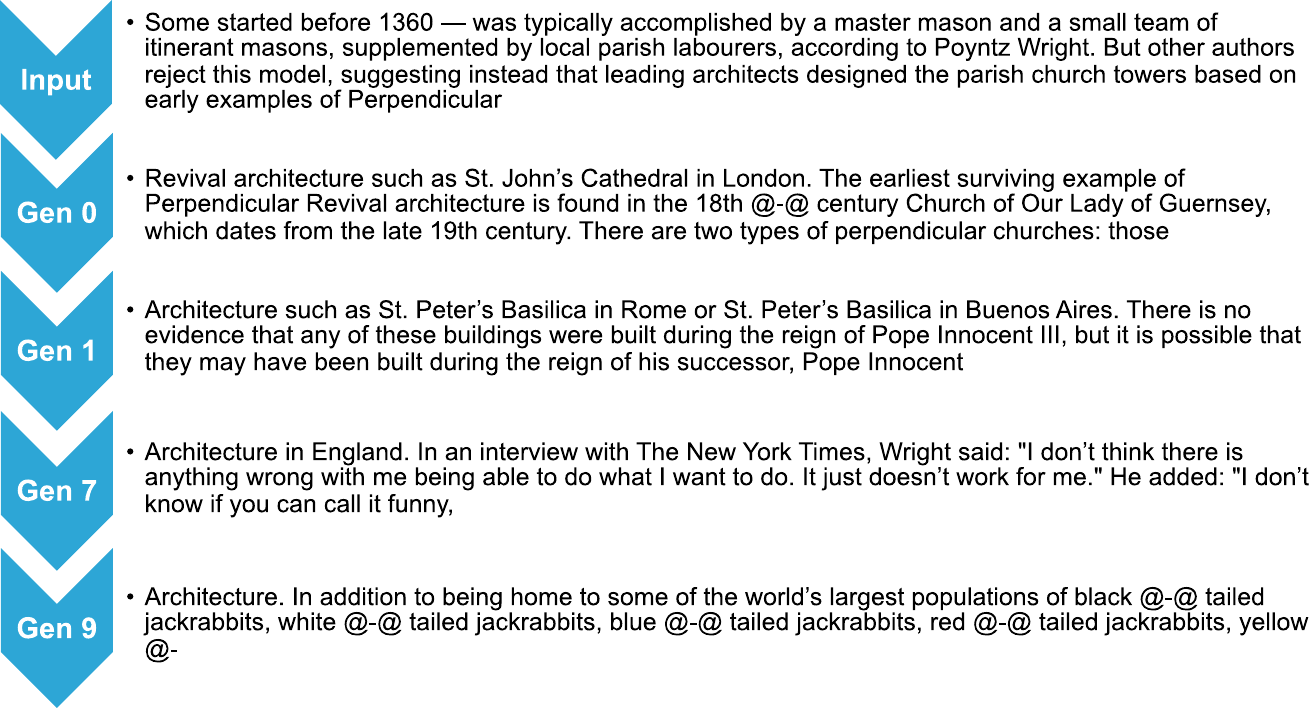}
    \caption{Synthetic texts from an autophagy loop from the study in \cite{RN4}, demonstrating a reduction in text variety. In later generations, the texts increasingly exhibit duplication.}
    \label{fig:language_quality}
\end{figure}

Empirical findings from both images shown in Figure \ref{fig:vision_quality}(a) and language data support these theoretical conclusions. For example, Alemohammad et al. \cite{RN5} observed that degradation in the StyleGAN2 model appears as cross-hatched artifacts, while Martínez et al. \cite{RN11, RN12} reported fuzziness and blurring in the DDIM model during autophagy. Bohacek et al. \cite{bohacek2023nepotistically} identified repetitive pixels and patterns in the Stable Diffusion Model. Similarly, in language models, Shumailov et al. \cite{RN4} found that the autophagy loop reduces output diversity, as shown in Figure \ref{fig:language_quality}. Additionally, Guo et al. \cite{guo2023curious} analysed diversity metrics across iterations, revealing a consistent decline in output diversity, particularly for tasks requiring high levels of creativity.

% \begin{figure}
%     \centering
%     \includegraphics[width=0.8\linewidth]{figs/image.pdf}
%     \caption{Synthetic images from an autophagy loop from paper \cite{RN5} and \cite{RN12}, showing the patterns of quality degradation using different generative AI backbone. }
%     \label{fig:vision_quality}
% \end{figure}

% \subsection{Mode Collapse and Model Collapse}
% The term "mode collapse" was used in the field of vision synthesis to describe a scenario where generative models, such as GANs, produce synthetic images with reduced variety, resulting in similar-looking images. Mode collapse occurs due to the GAN model's tendency to settle into local minima, influenced by the nature of discriminator loss. Even when mode collapse occurs in GANs, the model can still produce realistic-looking images, though with limited diversity.

% Based on this concept, subsequent research \cite{RN5} used the term "model collapse" to describe the decline in performance of generative models specifically caused by the addition of synthetic data into the training set. The paper \cite{RN5} further identified two specific types of model collapse: \textbf{early model collapse} and \textbf{late model collapse}. In early model collapse, the model begins losing information about the tails of the distribution, similar to GAN's mode collapse. In late model collapse, the model entangles different modes of the original distribution, converging to a distribution that bears little resemblance to the original, often with very small variance. 

\subsection{Fixed-Real Data Loop: Stability and Limitations}
\textcolor{black}{Introducing a fixed real dataset can create a \textbf{fixed-real data loop}, where AI-generated images are incrementally combined with real data. In each iteration, a fixed real dataset is incorporated. }

\textcolor{black}{Theoretically, Bertrand et al. \cite{bertrand2023stability} demonstrated the stability of iterative retraining when a fixed real dataset is preserved, i.e., if a fixed real data is observed, the quality degradation will become stable and not change. The study proposed a theoretical framework for iterative retraining of generative models, building upon the standard maximum likelihood objective. Fu et al. \cite{fu2024towards} also describes a theoretical guarantee on the stability and error bounds when training a model with a mixture of real and synthetic data. However, the stability in  \cite{bertrand2023stability} and \cite{fu2024towards} relies on unfeasible settings: (1) the initial generative model is sufficiently well-trained, and (2) each retraining iteration retains a high proportion of the original real data. }

\textcolor{black}{Gerstgrasser et al. \cite{gerstgrasser2024model} observed similar results using traceable linear models, where the test error stabilises at a finite upper bound as the number of iterations increases, rather than continuing to grow.  Nevertheless, this is still problematic: if the model has already collapsed into an unacceptable level of performance, even if the error ceases to worsen, the outcome remains unsatisfactory. }

\textcolor{black}{Empirically, keeping real data in the loop also slows the rate of model collapse \cite{gerstgrasser2024model,RN5,RN11,RN12,briesch2023large}, it does not solve the problem entirely.  }

\textcolor{black}{When real data is retained and synthetic data is scaled up, normal scaling laws break down. Typically, scaling laws predict performance improvements with more training data. However, mixing synthetic data with real data changes the data distribution, breaking the expected scaling law \cite{dohmatob2024tale}. }

\subsection{Fresh Data Loop: A Temporary Solution or a Real Fix?}
\textcolor{black}{A fresh data loop introduces new real data at each iteration. Studies \cite{RN5,briesch2023large} have shown that incorporating fresh real data can temporarily mitigate quality degradation. \cite{bohacek2023nepotistically} also demonstrated that adding fresh real data can partially heal the data loop.} 

\textcolor{black}{However, while the fresh data loop can \textbf{slow down} or \textbf{delay} degradation, it does not entirely resolve the underlying issue. The proportion of real data versus synthetic data becomes a critical factor: as the dataset grows, if the amount of synthetic data significantly outweighs the fresh real data being introduced \cite{azizi2023synthetic}, the model may still struggle to generalise effectively. This imbalance can still lead to performance stagnation or eventual collapse, albeit at a slower rate.}

\textcolor{black}{Moreover, the practical feasibility of consistently obtaining fresh real data is a challenge. In many real-world applications, collecting new high-quality data at every iteration is either expensive or impractical. Thus, while the fresh data loop offers a temporary buffer against the degenerative effects of synthetic data, it is not a complete solution.}

\subsection{The Diminished Role of Synthetic Data in Data Augmentation}
Before concerns arose about the AI autophagy loop, synthetic data was widely praised for its utility in augmenting training datasets, particularly for downstream tasks such as classification and image generation \cite{azizi2023synthetic,trabucco2023effective}. Synthetic data was recognised for improving sample variety and providing valuable pre-trained parameters. However, recent studies have raised doubts about its effectiveness.

\textcolor{black}{Ravuri and Vinyals \cite{ravuri2019classification} found that mixing generated samples with real data often degrades classifier accuracy. Although using low truncation values in GANs can improve accuracy when only a small amount of generated data is used, performance drops significantly when the proportion of synthetic data increases beyond a certain threshold, falling below the performance of models trained exclusively on real data.}

\textcolor{black}{Hataya et al. \cite{hataya2023will} empirically addressed this issue by simulating dataset contamination. They generated ImageNet-scale and COCO-scale datasets using state-of-the-art generative models and evaluated the impact of "contaminated" datasets on various tasks, including image classification and generation. Their findings showed that generated images negatively affect downstream performance, with the degree of degradation depending on the ratio of synthetic images to real images and the specific downstream task. Further analysis revealed that generated images also reduce robustness to out-of-distribution data, indicating that the use of synthetic data for augmentation poses a threat to the safety of AI models in open-world environments.}

\textcolor{black}{Chen et al. \cite{chen2024would} further highlighted the risks by focusing on dataset bias, showing that even well-generated images tend to amplify existing biases in the data.}

% Interestingly, simpler methods such as nearest-neighbor retrieval from existing datasets have outperformed sophisticated generative models like diffusion models in downstream tasks \cite{burg2023image}. This finding challenges the assumption that synthetic data can continually enhance model performance and suggests that more careful approaches are needed when using synthetic data in training pipelines.

\textcolor{black}{It is noteworthy noting that these findings have not accounted for autophagy loops, where synthetic data is reintroduced into the training cycle. Still, they offer a pessimistic view of the future impact of synthetic data on AI-based systems.}

\section{RQ2: What Technical Strategies Can Be Employed to Mitigate the Negative Consequences of AI Autophagy?}

\subsection{The Limitations of Cherry-Picking in Mitigating AI Autophagy}
A straightforward solution to mitigate model collapse is "cherry-picking"—removing low-quality or incorrect synthetic contents in generative models to ensure only high-quality, realistic ones are used in subsequent iterations of training. Techniques like the truncation trick in GANs \cite{RN18} and guidance parameters in diffusion models also help generate higher-quality images. Similarly, in language models, hallucinated or linguistically unacceptable content can be detected and filtered out \cite{guo2023curious} to maintain quality.

\textcolor{black}{However, these methods often worsen model collapse by reducing diversity, as noted in \cite{RN5}, as as shown in Figure \ref{fig:vision_quality} (b) and (c). A similar issue arises in language generation models. For example, Guo et al. \cite{guo2023curious} used a linguistic acceptability filter to remove noisy samples during each iteration, discarding the 20\% of synthetic data with the lowest acceptability scores before retraining the model. While this approach improves immediate data quality, it exacerbates model collapse by narrowing the model's generative variety. In contrast, Shumailov et al. \cite{RN4} introduced a penalty to reduce repetition, which helped mitigate model collapse. However, as a result of this penalty, models began generating lower-quality sentences in an effort to avoid repetition, ultimately reducing the overall performance of the model.}

\textcolor{black}{Researchers are also exploring feedback mechanisms to filter suboptimal samples \cite{feng2024beyond,ferbach2024self,gillman2024self}. Feng et al. \cite{feng2024beyond} showed that feedback-augmented synthetic data, such as pruning incorrect predictions or selecting the best guesses, can prevent model collapse, aligning with techniques like Reinforcement learning from human feedback (RLHF). Similarly, Ferbach et al. \cite{ferbach2024self} demonstrated that curating data based on a reward model maximises expected rewards, suggesting that increasing curated synthetic data optimises preferences for future models. However, these methods fail to fully resolve the inherent trade-off between quality and variety in the generative process, overlooking the challenge of designing a universal reward system for synthetic data selection in general-purpose AI.}

In addition to the previously discussed cherry-picking method, retaining real data in the training cycle has been both theoretically and empirically proven to mitigate the effects of model collapse \cite{RN5,briesch2023large}. \textcolor{black}{Additionally, \cite{alemohammad2024self} proposed using synthetic datasets to train an auxiliary model that adjusts the diffusion models in a negative direction, helping to prevent quality degradation during autophagy iterations. However, it is important to recognize that the success of these methods fundamentally relies on the accurate identification of synthetic data.} Thus:
\begin{tcolorbox}[breakable,notitle,boxrule=0pt,colback=gray!10,colframe=gray!20]
We must always be able to discern which parts of the training data are synthesised and which are human created.

\end{tcolorbox}

\subsection{Advantages and Caveats of Synthetic Content Watermarking}
Watermarking is one way to help users identify the synthetic contents \cite{wen2023tree, fernandez2023stable, bui2023trustmark, tancik2020stegastamp, RN40,RN41,RN42,RN43}. 

Watermarking synthetic images is a promising approach, already finding real-world applications in AI-driven image generation. In 2023, Google DeepMind and Google Cloud launched SynthID, a tool for watermarking and identifying AI-generated images. SynthID embeds an invisible digital watermark in an image's pixels, detectable for verification. It was released to select Vertex AI customers using Imagen, a text-to-image model. StableDiffusion also includes an invisible watermarking feature in its reference sampling script, using the Python package invisible-watermark with DCT and DWT algorithms.

Watermarking in language models has limited practical applications. In real-world use, users typically iteract with black-box LLMs, making it impractical to embed watermarks within the model’s generation process. Post-generation watermarking, which alters output texts, is less effective, often degrading text quality and remaining vulnerable to paraphrasing attacks \cite{RN44, RN45}.

\textcolor{black}{Robustness to attacks is a critical concern in watermarking methods. It is important to emphasise that our focus is not on extreme scenarios involving sophisticated, malicious adversarial attacks, but rather on ensuring that watermarks remain resilient against unintended attacks, perturbations and modifications that occur naturally, such as JPEG compression, cropping, and rephrasing of text. Robustness against these everyday alterations is a key quality metric in many technical studies \cite{RN22,RN34,RN44,RN41,RN43}. With proper stakeholder cooperation, current watermarking techniques are generally effective at resisting these simple distortions. However, this raises another practical question \cite{RN43}: if a human extensively rephrases AI-generated text, can it still be considered "AI-generated"? This issue is largely overlooked in the current literature, and there are no clear boundaries by which a reworded text remains AI-generated, nor the extent by which a reworded text used to update a generative model can contribute to the degradation and mode collapse described previously.}

Furthermore, effective watermark detection often depends on access to technical details that companies are typically unwilling to share, creating additional barriers to the effective implementation and reliability of watermarking systems.

% Similarly, watermarking is unlikely to prevent the misuse of NLP generative models unless uniformly implemented across all major large language models (LLMs). In real-world scenarios, users often interact with black-box LLMs, making it impractical to modify the model’s internal text generation or vocabulary distribution to embed watermarks. This significantly diminishes the effectiveness of LLM watermarking strategies. Post-generation watermarking methods, which involve altering the output texts after generation, are also less effective than integrated approaches, as they often compromise text quality and are highly vulnerable to paraphrasing attacks \cite{RN44, RN45}.

\subsection{Advantages and Caveats of Synthetic Content Detection}
In addition to watermarks, researchers are exploring passive detection (or forensics detection) algorithms \cite{passos2024review} based on the inherent differences between synthetic and human-generated content \cite{RN46, RN47, RN48,RN61,RN62,RN63,RN64,RN65}. These algorithms do not require direct interaction with the AI system that created the content.

Unlike imaging data, non-watermark-based detection techniques are prevalent for AI-generated language on commercial platforms. These techniques, such as GPTZero \cite{RN62}, Originality.ai \cite{RN74}, and Turnitin’s AI detection tools \cite{RN75}, are non-intrusive and do not compromise text quality.

However, a significant challenge for non-watermark based LLM detectors lies in their generalisability, or their capacity to recognise AI-generated texts from new models. As AI technology progresses, the distinction between human and AI-generated content is becoming increasingly blurred. For instance, in evaluations using a "challenge set" composed of high-quality synthetic and human-generated English texts, OpenAI’s text classifier correctly identified only 26\% of the generated content \cite{RN61}.

Concerns about the accuracy of synthetic text detectors go beyond precision, highlighting issues of explainability and fairness. Research \cite{RN76} shows that leading AI detectors often misclassify writings by non-native speakers as AI-generated. Some universities, like Vanderbilt \cite{RN77}, have stopped using these detectors because companies refuse to disclose their identification methods. 

%Vanderbilt University also raised concerns about external companies analyzing student data without clear privacy policies, underscoring the need for transparency and accountability in AI detector deployment.

\section{RQ3: Which Regulatory Strategies Can Be Employed to Address These Negative Consequences?}
\textcolor{black}{Section 3 explores technical solutions to tackle the problem of AI autophagy, emphasizing the importance of marking and detecting synthetic content. This approach is identified as the most practical and straightforward method to address the issue. Integrating technical and regulatory efforts is crucial to ensure the success of these technological approaches.} 

\subsection{Problematic Data Acquisition and Dissemination Strategies}
\textcolor{black}{Current policies of online generative platform companies have failed to significantly address the AI autophagy loop. This failure is evident in their unregulated absorption of online data without implementing strategies to filter synthetic content \cite{openAI2024dalle, openAI2024safety, stabilityAI2024,googleAI2024}. While these companies have adopted responsible AI measures to limit the inclusion and dissemination of harmful or unethical content \cite{googleAI2024, openAI2024safety}, they have not specifically addressed the issue of synthetic content. Current strategies primarily focus on reducing explicit, violent, or biased content during distribution, rather than marking and filtering synthetic synthetic content from training processes, leaving a significant gap in combating the AI autophagy loop.}

\textcolor{black}{Although solid evidence directly linking these problematic policies to negative impacts is limited, some indications of their effects include performance degradation in models such as GPT-3.5 and GPT-4 \cite{RN7}. These models exhibited increased formatting errors in code generation from March to June 2023  \cite{RN7}. While the exact causes are not fully clear, it is speculated that the pollution of training datasets with AI-generated data could be a contributing factor \cite{sanusi2023}.}

\subsection{Relevant Regulations for the Dissemination Process}
\textcolor{black}{Combating the AI autophagy loop requires government regulations to ensure that synthetic content is identifiable and detectable, facilitating its filtering from training datasets. This transparency is essential not only for addressing the AI autophagy loop but also for broader concerns such as preventing the spread of misinformation, protecting intellectual property, and maintaining trust in digital media.}

\textbf{China}: China has issued the \textit{Interim Measures for the Management of Generative AI Services}, requiring providers to clearly mark generated content, such as images and videos, to prevent public confusion (Article 12) \cite{RN81}. Similarly, the \textit{Artificial Intelligence Act of 2021} mandates that users of AI systems generating realistic content, such as deepfakes, must label such content as artificially generated or manipulated.

\textbf{European Union (EU)}: \textcolor{black}{The \textit{EU AI Act (2024)} categorises AI systems based on their risk levels \cite{madiega2021ai,EU2024}. AI systems must comply with information and transparency requirements. Users must be informed when they are interacting with AI, such as chatbots. Deployers of AI systems that generate or manipulate images, audio, or video content, including deepfakes, are required to disclose that the content has been artificially generated or manipulated, except in limited cases such as for the prevention of criminal offenses. Providers of AI systems generating large quantities of synthetic content must implement reliable, interoperable, and robust techniques, such as watermarks, to mark and identify AI-generated or manipulated content.}

\textbf{United States (US)}: \textcolor{black}{The US has introduced a presidential regulation focused on addressing AI risks. The regulation calls for the development of effective labeling and content provenance mechanisms, enabling Americans to determine when content has been generated or manipulated by AI \cite{nist2023safe}. These actions are intended to address AI's risks while preserving its potential benefits.}

\textcolor{black}{While it is still early to fully assess the impact of these regulations, they have already begun to influence the global conversation on AI governance and accountability \cite{diaz2023connecting}. Regulations are crucial for controlling the misuse of synthetic data, but relying solely on the integrity of synthetic data disseminators is impractical \cite{RN84}. Without effective detection methods for identifying synthetic data, ensuring compliance with these laws remains a significant challenge.}

\section{Conclusions and Outlook}

In this paper, we revisit the concept of AI autophagy, where generative models iteratively consume their own synthetic outputs. Our perspective covered theoretical insights, experimental findings, and potential technical and regulatory solutions to address this phenomenon. Below, we summarise the findings, structured around the original questions posed in the introduction:

\begin{itemize}
\item \textbf{RQ1: What are the consequences of uncontrolled autophagy with synthetic contents?}
\end{itemize}

\textcolor{black}{When AI models consume their own outputs—a phenomenon known as AI autophagy—the data distribution becomes increasingly distorted \cite{dohmatob2024tale,RN4}, leading to significant performance degradation over time. Iterative training on synthetic data causes a gradual decline in both quality and diversity of generated outputs, as demonstrated by both theoretical models, including Gaussian \cite{RN4,RN5,bertrand2023stability}, probabilistic \cite{bertrand2023stability,fu2024towards}, and linear \cite{dohmatob2024model,gerstgrasser2024model} frameworks, and empirical findings \cite{RN4,RN5,RN11,RN12,bohacek2023nepotistically,guo2023curious,dohmatob2024tale}.}

\begin{figure}[h!]
    \centering
    \includegraphics[width=\linewidth]{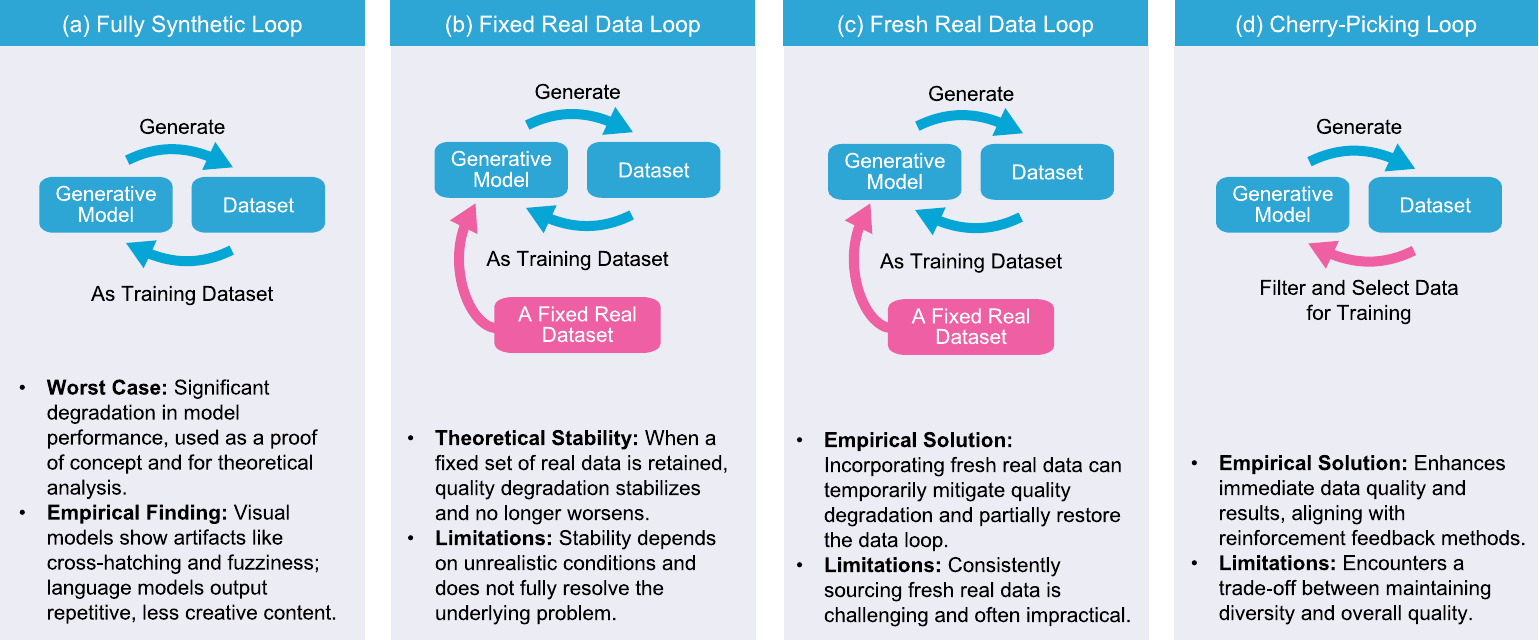}
    \caption{Different loops in AI autophagy.}
    \label{fig:loops}
\end{figure}

\textcolor{black}{Incorporating fixed-real data \cite{bertrand2023stability,gerstgrasser2024model} or introducing fresh data \cite{RN5}, as shown in Figure \ref{fig:loops}, into training cycles offers only temporary relief. While these strategies can slow the rate of collapse, they do not address the fundamental issue, as maintaining a balance between synthetic and real data is challenging, particularly in real-world scenarios where high-quality real data is often scarce. }

\begin{itemize}
\item \textbf{RQ2: What technical strategies can be employed to mitigate these negative consequences?}
\end{itemize}

\textcolor{black}{While strategies like incorporating real data into the training process and filtering out low-quality synthetic data can mitigate these issues, these approaches are constrained by practical limitations. Incorporating real data requires consistent sourcing, which is often impractical, and filtering strategies can inadvertently reduce data diversity, exacerbating long-term model collapse. Moreover, the effectiveness of these mitigation methods fundamentally depends on accurately distinguishing between real and synthetic data—a task that grows increasingly complex as synthetic data becomes more sophisticated. }

Common techniques for distinguishing synthetic content include watermarking and detection methods. While watermarking is cost-effective, it can impact content quality and is not universally applicable. Detection methods avoid these drawbacks but struggle with generalisesation and transparency. Implementing widespread protection requires collaboration from regulatory authorities and commercial entities, but universal strategies risk uniform attacks. Additionally, concerns about the accuracy, explainability \cite{RN77} and fairness \cite{RN76} of synthetic content detectors persist.
% \begin{figure}[h!]
%     \centering
%     \includegraphics[width=\linewidth]{figs/methods.pdf}
%     \caption{Pros and cons of technical solutions to data pollution caused by AI autophagy.}
%     \label{fig:technical}
% \end{figure}

\begin{itemize}
\item \textbf{RQ3: Which regulatory strategies can be employed to mitigate these negative consequences?}
\end{itemize}
\textcolor{black}{Current policies implemented by generative AI companies are insufficient, highlighting the need for stronger collaboration between regulatory bodies and corporations. Regulations must work in tandem with industry efforts to develop effective techniques for identifying AI-generated content, which can then be filtered out from training datasets. This approach would not only address the AI autophagy loop but also mitigate broader concerns, such as the spread of misinformation, intellectual property violations, and the erosion of trust in digital content.}

Current regulatory strategies proposed include mandating the marking of generated content to prevent public confusion, requiring real-name verification for content creators to enhance traceability, and enforcing clear labeling of AI-generated content, particularly deep fakes. However, the effectiveness of these strategies relies on the development of reliable detection methods for AI-generated content, demonstrating a need for a collaborative effort among technology companies, regulatory agencies, and civil society to forge effective governance strategies, safeguarding the web's content integrity.
\begin{figure}[h!]
    \centering
    \includegraphics[width=\linewidth]{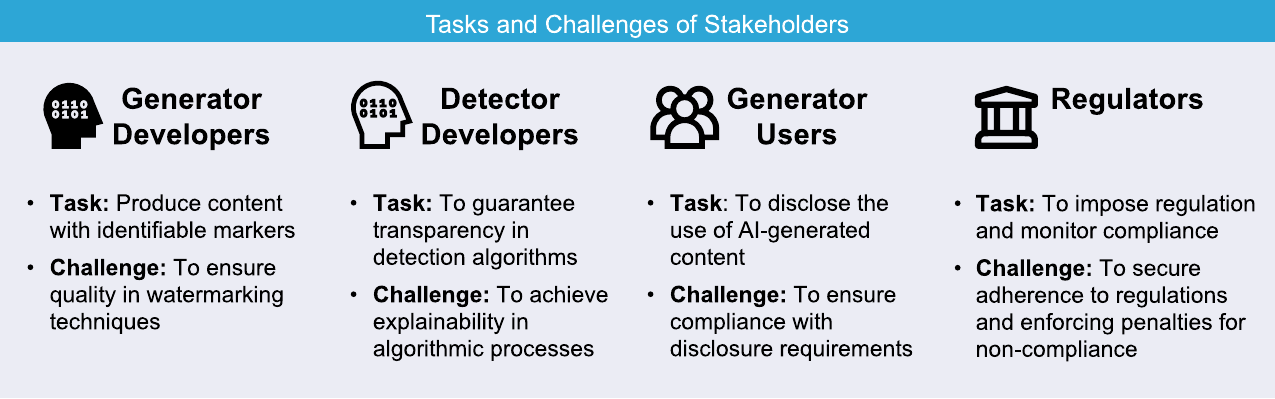}
    \caption{Overview of tasks and challenges faced by different stakeholders in the context of preventing data pollution by generative models.}
    \label{fig:conclusion}
\end{figure}

% \subsection{Future Implications and Recommendations }
\textcolor{black}{A key takeaway from our paper is that no single technical solution stands out as a definitive answer to addressing the AI autophagy issue. While maintaining a certain amount of real data and filtering out high quality synthetic data for training serves as a temporary fix, its effectiveness still relies on our ability to distinguish between synthetic and real data accurately.}

As a summarisation of this work, we present in Figure \ref{fig:conclusion} an overview of the roles and corresponding challenges faced by various stakeholders in the context of AI-generated content. \textcolor{black}{ \textbf{Generator developers} are individuals or organizations, including labs or large companies, that create and refine deep generative models. Their key task is to embed identifiable markers, like watermarks, in AI-generated content to distinguish it from human-made material, without degrading quality—requiring sophisticated techniques. \textbf{Detector developers}, which may include governments and authorities, focus on building tools to detect AI-generated content. Their role is vital for verifying authenticity, and they must ensure detection systems are transparent and trustworthy. Their challenge is balancing explainability with maintaining the accuracy and robustness of their systems. \textbf{Generator users}—such as content creators, businesses, and researchers—use AI-generated content and are responsible for disclosing its use. They play a crucial role in ensuring transparency to prevent misinformation, but complying with disclosure requirements can be difficult, especially if regulations are not strictly enforced. \textbf{Regulators}—government agencies, international bodies, or industry groups—create and enforce rules for AI-generated content. They are responsible for mandating watermarking, detection, and disclosure, and for protecting the public from risks like misinformation. Their challenge is keeping regulations enforceable in a rapidly changing AI landscape.}

With the responsibilities and challenges clearly identified among these stakeholders, we can focus on building a comprehensive framework that enables collaboration across sectors to prevent the risks associated with AI-generated content. By working together, generator developers, detector developers, generator users, and regulators can address the shortcomings of current technical solutions, such as watermarking and detection tools, and ensure that these technologies are robust and scalable in real-world applications. This collaborative framework will also strengthen the enforcement of regulations, helping to mitigate issues like misinformation, data pollution, and non-compliance, which are critical in a rapidly evolving AI landscape.

\section{Ethical and Societal Considerations}
The aim of this study is to assess the potential adverse effects of autophagy loops in generative models, particularly the impact of synthetic data proliferation across the web. These findings highlight the risks associated with unchecked AI-generated content, such as degradation in model quality, misinformation, and potential harm to data integrity in various fields.

It is important to emphasize that this concern does not imply endorsement of practices such as large-scale web scraping for training generative AI models. The ethical implications of data sourcing, especially without consent, pose serious challenges to privacy and data ownership, and should be carefully regulated.

The intended use of the findings presented in this paper is to encourage collaborative efforts among generator developers, detector developers, users, and regulators to create frameworks that promote transparency, accountability, and responsible AI use. It is our hope that these solutions can help mitigate risks such as data pollution and ensure AI-generated content is used ethically and for the benefit of society.

This work also acknowledges that while technological solutions, such as watermarking and detection systems, are promising, they are not without limitations. Ensuring compliance with ethical standards in real-world applications requires robust enforcement mechanisms, cooperation among stakeholders, and ongoing dialogue with policymakers and civil society.

\section{Acknowledgements}
This study was supported in part by the ERC IMI (101005122), the H2020 (952172), the MRC (MC\/PC\/21013), the Royal Society (IEC\textbackslash NSFC\textbackslash 211235), the NVIDIA Academic Hardware Grant Program, the SABER project supported by Boehringer Ingelheim Ltd, the UKRI Future Leaders Fellowship (MR\/V023799\/1), and Wellcome Leap’s Dynamic Resilience programme, jointly funded with Temasek Trust. This work was made possible through the support of the ARC project team at The Alan Turing Institute, under the project designation ARC-001. J. Del Ser acknowledges funding support from the Basque Government through EMAITEK/ELKARTEK grants, as well as the consolidated research group MATHMODE (IT1456-22).

Correspondence should be sent to G. Yang (\texttt{g.yang@imperial.ac.uk}).

\section{Competing interest statement}
\textcolor{black}{All authors declare no competing interests.}

\bibliography{sn-bibliography}% common bib file
%% if required, the content of .bbl file can be included here once bbl is generated
%%\input sn-article.bbl

\end{document}